\pdfoutput=1
\documentclass[11pt]{article}

\usepackage[final]{acl}

\usepackage{times}
\usepackage{latexsym}
\usepackage{longtable}
\usepackage[T1]{fontenc}

\usepackage[utf8]{inputenc}

\usepackage{microtype}

\usepackage{inconsolata}

\usepackage{graphicx}

\usepackage{tipa}

\newcommand{\clicsfour}{CLICS~4}

\title{Advancing the Database of Cross-Linguistic Colexifications with New Workflows and Data}

\author{
 \textbf{Annika Tjuka\textsuperscript{1}},
 \textbf{Robert Forkel\textsuperscript{1,2}},
 \textbf{Christoph Rzymski\textsuperscript{1}},
 \textbf{Johann-Mattis List\textsuperscript{2,1}}
\\
\\
 \textsuperscript{1}Max Planck Institute for Evolutionary Anthropology, Leipzig, Germany\\
 \textsuperscript{2}Chair of Multilingual Computational Linguistics, University of Passau, Passau, Germany
\\
 \small{
   \textbf{Correspondence:} \href{mailto:annika_tjuka@eva.mpg.de}{annika\_tjuka@eva.mpg.de} and \href{mailto:mattis.list@uni-passau.de}{mattis.list@uni-passau.de}
 }
}

\begin{document}
\maketitle
\begin{abstract}
Lexical resources are crucial for cross-linguistic analysis and can provide new insights into computational models for natural language learning. Here, we present an advanced database for comparative studies of words with multiple meanings, a phenomenon known as colexification. The new version includes improvements in the handling, selection and presentation of the data. We compare the new database with previous versions and find that our improvements provide a more balanced sample covering more language families worldwide, with enhanced data quality, given that all word forms are provided in phonetic transcription. We conclude that the new Database of Cross-Linguistic Colexifications has the potential to inspire exciting new studies that link cross-linguistic data to open questions in linguistic typology, historical linguistics, psycholinguistics, and computational linguistics.
\end{abstract}

\section{Introduction}

The \emph{Database of Cross-Linguistic  Colexifications} \citep[CLICS, \href{https://clics.clld.org}{https://clics.clld.org},][]{Rzymski2020} offers detailed data on the distribution and frequency of \emph{colexifications} across several thousand languages. Colexification is a cover term that unifies the notions of polysemy, homophony, and underspecification, referring to cases where a single word form in a given language expresses multiple senses \citep{Francois2008}.
For example, Vietnamese \textit{xanh} refers to `blue' and `green' at the same time, German \emph{b\"ose} means both `angry' and `evil', or English \textit{ear} refers to a part of the body or a part of a grain. The different examples represent words with multiple senses and can be labeled as underspecification (Vietnamese), polysemy (German), or homophony (English), but they can also be taken together as examples of the phenomenon of colexification.

 
CLICS has built on this idea by collecting data from multilingual word lists that were unified with respect to the semantic glosses by which words across different languages are elicited. From these word lists, colexifications were automatically extracted, forming a large \emph{colexification network} \citep{List2013} that can be investigated interactively \citep{Mayer2014}. The database has improved concerning the workflow by which data are aggregated and in terms of the number of datasets underlying the database (4 datasets in Version 1.0, \citealt{List2014}, 15 datasets in Version 2.0, \citealt{List2018a}, 30 datasets in Version 3.0 \citealt{Rzymski2020}). In its current form, the CLICS database is characterized by three major features. First, CLICS \emph{aggregates} data from existing standardized datasets, rather than curating data directly. Second, CLICS offers its data in both \emph{machine- and human-readable form}, allowing scholars to access the data in computational workflows as well as through the web interface. Third, CLICS is \emph{open}, and both the individual data and the source code are published with permissive licenses, allowing scholars not only to investigate the database, but also to extend it with additional content or methods.

\begin{figure*}[tb]
\centering
\includegraphics[width=\textwidth]{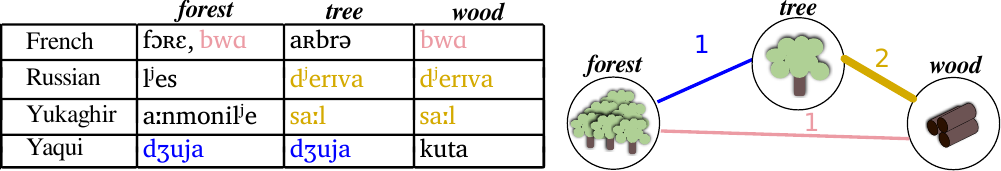}
\caption{Cross-linguistic colexifications (left) and cross-linguistic colexification network (right). The figure illustrates how colexification networks can be reconstructed from cross-linguistic colexification data, using information obtained from the CLICS database (Version 3.0, \citealt{Rzymski2020}).}
\label{fig:clics}
\end{figure*}
 
Given that five years have passed since the last official release of CLICS and that new relevant datasets have been published during this time, mainly as part of Lexibank, a large repository for standardized multilingual word lists \citep[\href{https://lexibank.clld.org}{https://lexibank.clld.org},][]{List2022,Blum2025b}, it is time to improve the database even further. Taking advantage of the fact that CLICS is open and free to modification, we therefore present an updated version of the CLICS database, which we named \clicsfour{} for convenience. \clicsfour{} not only increases the underlying data, but also addresses three major shortcomings of the previous versions of CLICS by improving (1)~the handling of concepts (§~\ref{sec:3.2}), (2)~the selection of languages to be included in the colexification database (§~\ref{sec:3.3}), and (3)~the general representation of data (§~\ref{sec:3.4}). In the following, we will present previous studies devoted to cross-linguistic colexifications (§~\ref{sec:2}), discuss the improvements in more detail (§~\ref{sec:3}), illustrate their consequences for \clicsfour{} (§~\ref{sec:4}), and reflect on the future of cross-linguistic colexification data (§~\ref{sec:5}).

\section{Background}
\subsection{Cross-Linguistic Colexifications}\label{sec:2}

Not long after \citet{Francois2008} had first introduced the term \emph{colexification} along with initial ideas on how the phenomenon could be analyzed using cross-linguistic data, typologists quickly adopted the term and the technique to study lexical semantics both globally and in certain linguistic areas. Two major reasons contributed to the popularity of the term and the technique. 

First, polysemy and homophony are notoriously difficult to distinguish, specifically when analyzing languages whose history is less well known. While scholars sometimes distinguish both relations by degree of semantic similarity, arguing that homophonous words show greater divergence in meaning than polysemous words \citep[7]{Leivada2021}, the original distinction between polysemy and homophony is strictly diachronic. Thus, they reflect two distinct processes of language change: polysemy is the result of semantic change, while homophony is the result of a merger of originally distinct word forms due to sound change (\citealt[12f]{Sperber1923}, \citealt[11]{Apresjan1974}). However, in the minds of speakers, the history of the words does not play a major role. Speakers seem to show some general awareness that some words have multiple senses that are closely related to each other, whereas other words with distinct senses merely sound alike \citep[20f]{Enfield2015}. While it may seem useful to distinguish polysemy and homophony in theory, the distinction of the two relations in practice is difficult to make. Omitting the explicit distinction between the two forms of lexical ambiguity allowed scholars to assemble data in an unbiased and efficient way. Scholars could accumulate colexification data for their areas of interest without having to discuss the consequences of impractical terminology. Instead of deciding whether the findings would reflect polysemy or homophony, scholars could let the data decide, given that polysemy often largely exceeds homophony. 

Second, scholars began to explore the benefits of modeling cross-linguistic colexifications with the help of network approaches \citep{Cysouw2010}. This not only led to clear visualizations of semantic similarities that could be observed across languages, but also opened up new possibilities for the analysis of cross-linguistic polysemy using network approaches \citep{List2013} and the introduction of interactive techniques for data visualization and exploration, which later became a core component of the CLICS database \citep{Mayer2014}. Figure \ref{fig:clics} illustrates how colexification networks can be constructed from colexification data, using data from CLICS~3 \citep{Rzymski2020}.

Due to this approach, which facilitates the collection of data and offers new ways to analyze the data through inspection and computation, cross-linguistic colexifications have become an integral part of lexical typology, with a multitude of applications in studies on semantic similarity. CLICS offered the first and largest collection of cross-linguistic colexifications and was used in several studies, examining a large number of topics, ranging from investigations on genealogical language relations \citep{Blevins2021,Blum2024d} and linguistic areas \citep{Gast2019}, via analyses of particular semantic domains \citep{Jackson2019,DiNatale2021,Brochhagen2022,Tjuka2024}, up to initial applications in computational linguistics \citep{Bao2021,Bao2022} and communication science \citep{Bradford2022}. In addition, CLICS is now regularly consulted in typological studies that explore particular phenomena in detail, allowing authors to contrast their findings with their insights on a specific group of languages \citep{Sjoeberg2023,Souag2022,Schapper2019,Schapper2022a}.

To summarize, cross-linguistic colexifications
and cross-linguistic colexification networks have
become a crucial tool in comparative linguistics.
The application of cross-linguistic colexification analysis is not restricted to lexical typology,
but provides interesting insights into additional fields of linguistics
and beyond, including historical linguistics, areal linguistics, computational
semantics, and human cognition.

\subsection{Data Aggregation and Analysis in CLICS}\label{sec:2.2}
The integral part of the Database of Cross-Linguistic Colexifications is the workflow by which data are aggregated from individual datasets and later analyzed to create a colexification network.
To be able to aggregate data from different resources, datasets must be standardized.
Standardization is achieved with the help of
Cross-Linguistic Data Formats \citep[CLDF, \href{https://cldf.clld.org}{https://cldf.clld.org},][]{Forkel2018}, an initiative that builds on the CSVW standard for tabular data on the web \citep[\href{https://csvw.org}{https://csvw.org},][]{CSVW}, but extends CSVW with semantics relevant to comparative linguistics. A CLDF dataset is a collection of CSV files linked via a JSON file that stores the metadata, providing information on how the CSV files should be interpreted computationally and what values are shared across the files. Thus, a CLDF dataset is a small relational database with specific semantics that link the data with additional data from outside. 

The most important external datasets that CLICS links to are three \emph{reference catalogs}: \emph{Glottolog}, \emph{Concepticon}, and \emph{CLTS}. Glottolog \citep[\href{https://glottolog.org}{https://glottolog.org},][]{Glottolog} offers basic information on language varieties, including information on language classification, geolocations, and the documentation status of individual languages. Concepticon \citep[\href{https://concepticon.clld.org}{https://concepticon.clld.org},][]{Concepticon} offers a collection of basic senses that are expressed across multilingual word lists. Senses are provided in the form of \emph{concept sets} that are linked across several hundred \emph{concept lists} that have been annotated by the Concepticon team in the past decade \citep[for details on the curation process, see][]{Tjuka2023}. CLTS \citep[\href{https://clts.clld.org}{https://clts.clld.org},][]{CLTS} is a reference catalog for \emph{Cross-Linguistic Transcription Systems} that standardizes phonetic transcriptions by advocating a subset of the International Phonetic Alphabet \citep{IPA1999} that is represented in the form of distinctive features (for details, see \citealt{Anderson2018} and \citealt{Rubehn2024a}). The conversion of individual datasets to the CLDF standard is supported by dedicated Python libraries (most importantly the \texttt{CLDFBench} packages, \citealt{Forkel2020}) that help to check the overall consistency of the data.
 
From a collection of CLDF datasets, the CLICS aggregation workflow iterates over the datasets and assembles cross-linguistic colexifications for each language variety. Here, CLICS uses an efficient method that avoids comparing $n$ words in one language against $n$ words in the same language, but rather identifies colexifications from tuples, consisting of a word form and its corresponding sense (a \textit{concept set} in the Concepticon catalog), with the help of hash tables \citep{List2022TBLOG06}. In other words, the method iterates over all words in a dataset only once, instead of comparing all words against each other, which would result in large computation times.
 
Having created a large colexification network of all CLDF datasets, the CLICS workflow analyzes this data further by computing \emph{communities}, that is, partitions of nodes in a graph that show more connections to each other than to other nodes outside of the partition \citep[8577]{Newman2006b}. Communities are inferred with the help of the Infomap algorithm \citep{Rosvall2008} and are used to structure the web application, by allowing users to inspect either entire communities or individual subsets of the data.
The methods for data aggregation and analysis are freely accessible and can be easily applied by scholars to create their analyses of subsets of the data in CLICS or by extending the CLICS collection further, as illustrated, for example, by \citet[][]{Tjuka2024b}.

\subsection{Shortcomings of CLICS}\label{sec:2.3}
Although the CLICS database serves as a main provider of cross-linguistic information on colexifications, CLICS~3 showed four major shortcomings that need to be addressed to ensure that future findings based on the data are solid and reliable. 
 
The first shortcoming relates to the data underlying CLICS. While data from 30 datasets were aggregated in Version 3.0 \citep{Rzymski2020}, many more datasets have recently been made available via the Lexibank repository \citep{List2022}. Improving the database by increasing the number of datasets is thus one of the most urgent tasks that should be addressed in an updated version. 
 
The second shortcoming relates to the treatment of concepts in the database. CLICS~3 used a rather naïve approach by taking concept sets provided by the Concepticon reference catalog at face value, without considering their interdependencies. Concepticon has several concept sets that appear in a hierarchical relation to other concept sets, mostly reflecting cases of underspecification, such as the concept set \href{https://concepticon.clld.org/parameters/2382}{\textsc{blue or green}}, expressed in the Vietnamese word \textit{xanh}. The colexification inference workflow in CLICS~3 treats the colexification of \href{https://concepticon.clld.org/parameters/837}{\textsc{blue}} and \href{https://concepticon.clld.org/parameters/1425}{\textsc{green}} expressed by the word \textit{xanh} as a single concept. However, this omits valuable colexification information.

Third, CLICS~3 provided information from more than 3,000 language varieties. However, a closer look at the data showed that only a small proportion of the included languages met the requirement set by the editors of CLICS~3 to provide elicitation glosses for at least 250 concepts. For CLICS~3, the authors instead selected 30 datasets that were officially compiled from concept lists with 250 or more items. The resulting word lists for individual languages, however, were often scarce and a larger number of the languages did not meet the originally stated coverage criterion. 

Fourth, CLICS~3 offered the colexification network exclusively in the form of a GML file. Although GML is a common format for the encoding of graphs \citep{Himsolt2010}, accepted by many software tools, including \texttt{igraph} \citep[\href{https://igraph.org}{https://igraph.org,}][]{Csardi2006}, \texttt{NetworkX} \citep[\href{https://networkx.org/}{https://networkx.org/,}][]{Hagberg2008}, and \texttt{Cytoscape} \citep[\href{http://cytoscape.org/}{http://cytoscape.org/,}][]{Smoot2011}, the format is not well-suited to share the extensive data on colexification patterns computed by CLICS~3. As a result, more transparent data formats for handling colexification data and colexification networks are needed to represent the results of the CLICS workflow in detail.

With the increasing use of CLICS~3, it is time to tackle these four points of criticism. In this study, we address these shortcomings by creating an updated version of CLICS that substantially increases the amount of data, improves the handling of concepts, corrects for the bias in language and concept selection, and makes the data representation more transparent.

\section{Materials and Methods}\label{sec:3}
In the following, we will introduce all necessary steps that lead to the creation of our modified \clicsfour{} database. We followed the established workflow for data aggregation used in CLICS~3 to some extent (§~\ref{sec:2.2}). However, we present a drastic increase of data based on standardized datasets (§~\ref{sec:3.1}), introduce an improved handling of concepts during data aggregation (§~\ref{sec:3.2}), refine the selection of languages and concepts (§~\ref{sec:3.3}), and make the representation of the colexification data more transparent (§~\ref{sec:3.4}). 

\subsection{Data Basis}\label{sec:3.1}

CLICS~3 was based on 30 datasets available in CLDF. Many more datasets have since been published as part of the Lexibank repository, which was first published in 2022 \citep{List2022} as Lexibank 1 and curates data from 100 different datasets of different sizes. Of those 52 Lexibank datasets were suitable for inclusion in CLICS, because they were based on concept lists that contain 250 or more items (this criterion was used to build CLICS~3, \citealt{Rzymski2020}). The newest version, Lexibank~2, offers data for 134 different datasets that are all phonetically transcribed \citep{Blum2025b}. For our enhanced version of CLICS, we identified 95 suitable datasets. These datasets are listed in the supplementary material accompanying this study.  

The datasets include cross-linguistics studies of specific language groups \citep[e.g.,][]{Bowern2012,Bodt2019} and global collections such as the Intercontinental Dictionary Series \citep[IDS, \href{https://ids.clld.org}{https://ids.clld.org},][]{Key2023} or the World Loanword Database \citep[\href{https:/wold.clld.org}{https://wold.clld.org},][]{Haspelmath2009}. The latter datasets were not originally provided together with phonetic transcriptions, but recent studies have added them (see \citealt{Miller2020} for WOLD and \citealt{List2023a} and \citealt{Miller2024TBLOG09} for IDS).

\subsection{Concept Handling}\label{sec:3.2}
The colexifications in CLICS result from comparing words mapped to the standardized concept sets in Concepticon \citep{List2016,Tjuka2023}. The consequent mapping of the elicitation glosses in individual datasets to the Concepticon reference catalog has been one of the most important factors that allowed for the growth of CLICS: Version 1.0 \citep{List2014} containing 221 language varieties and 1,280 concepts, Version~2.0 \citep{List2018a} containing 1,220 language varieties and 2,487 concepts, and Version 3.0 \citep{Rzymski2020} containing 3,156 language varieties and 2,906 concepts. 
However, through the mapping of the datasets to the Concepticon, a bias for a certain number of concepts that exhibit hierarchical relations to other concepts was introduced. 
 
Already with its first launch \citep{List2016}, Conception has allowed for the definition of broad concepts that are expressed as such only in specific languages or specific linguistic areas. As an example, consider the concept sets \href{https://concepticon.clld.org/parameters/2121}{\textsc{arm or hand}} and \href{https://concepticon.clld.org/parameters/2098}{\textsc{foot or leg}}. These concept sets are expressed by individual word forms in languages such as Vietnamese \textit{tay}, referring to `arm' or `hand', or Russian \emph{noga}, referring to `foot' or `leg'. However, many languages distinguish them further, using individual words for \href{https://concepticon.clld.org/parameters/1673}{\textsc{arm}}, \href{https://concepticon.clld.org/parameters/1277}{\textsc{hand}}, \href{https://concepticon.clld.org/parameters/1301}{\textsc{foot}}, and \href{https://concepticon.clld.org/parameters/1297}{\textsc{leg}}, respectively. 
 
Some lists in Concepticon have a linguistic area or language family as a target. Thus, the introduction of underspecified concept sets, such as \href{https://concepticon.clld.org/parameters/2121}{\textsc{ARM OR HAND}} or \href{https://concepticon.clld.org/parameters/2098}{\textsc{FOOT OR LEG}} was important, because linguists reporting on Slavic languages or particular languages in South-East Asia do not elicit both \href{https://concepticon.clld.org/parameters/1673}{\textsc{arm}} and \href{https://concepticon.clld.org/parameters/1277}{\textsc{hand}}, if they know that these are always colexified in the languages under study.
However, this kind of lexical underspecification, as we encounter it in the lexicons of Vietnamese and Russian, is one of the typical reasons for colexifications. Therefore, it is important to list such cases as true colexifications of \href{https://concepticon.clld.org/parameters/1673}{\textsc{arm}} and \href{https://concepticon.clld.org/parameters/1277}{\textsc{hand}}, as well as \href{https://concepticon.clld.org/parameters/1301}{\textsc{foot}} and \href{https://concepticon.clld.org/parameters/1297}{\textsc{leg}}. The original aggregation technique used by CLICS ignores these cases. As a result, important colexification information for a large number of languages is lost.


 
In our updated version \clicsfour{}, we account for underspecification directly, by defining a list of 85 concept sets that exhibit underspecification along with the more specific target concepts that they cover. While most of these underspecified concept sets can be represented by two concept sets, some are represented by more than two (specifically kinship terms like \href{https://concepticon.clld.org/parameters/1263}{\textsc{sister}}, which has four counterparts: \href{https://concepticon.clld.org/parameters/2420}{\textsc{younger sister (of man)}}, \href{https://concepticon.clld.org/parameters/2421}{\textsc{younger sister (of woman)}}, \href{https://concepticon.clld.org/parameters/2418}{\textsc{older sister (of man)}}, and \href{https://concepticon.clld.org/parameters/2419}{\textsc{older sister (of woman)}}). In addition, we decided to replace some concept sets with a too broad or too narrow definition by more common concept sets (e.g. replacing \href{https://concepticon.clld.org/parameters/2125}{\textsc{stone or rock}} by \href{https://concepticon.clld.org/parameters/857}{\textsc{stone}} because \href{https://concepticon.clld.org/parameters/668}{\textsc{rock}} did not occur in the data). 
 
When encountering words that are mapped to these concepts during the initial iteration over all word lists in the data, the respective words are multiplied and each of the words is mapped to the specific concept sets covered by the underspecified concept sets. Word forms that are artificially multiplied in this form are marked in the resulting dataset by providing information on the original concept set. In total, we identify 85 underspecified concept sets in Concepticon that are relevant for the data in our modified version of CLICS. Of the 1,445,845 word forms in \clicsfour{}, 107,921 word forms result from this refinement procedure. A detailed list of the concept replacements can be found in Appendix~\ref{sec:replacements}.


\subsection{Language and Concept Selection}\label{sec:3.3}
CLICS~3 included data from 3,156 language varieties. The criterion for including a given word list in the database was the size of the concept list underlying the respective dataset. The idea was to include only those languages with word forms for 250 or more concepts. However, since the editors of CLICS~3 only checked the size of the concept lists at the level of entire datasets, the CLICS~3 data contained a large number of language varieties with much fewer than 250 concepts covered. When discarding those varieties that contain fewer than 250 word forms, only 1,674 varieties remain.
 
After detecting this problem when reviewing individual datasets in CLICS~3, we decided to modify the criterion for the selection of languages in three ways. First, instead of setting the threshold to 250 words per language, we lowered it to 180 words, accounting for the fact that almost half of the languages in CLICS~3 would not pass this threshold. The threshold was chosen because we noticed that there were many datasets with 200 words or fewer. For many languages, only versions of the Swadesh list with 200 concepts \citep{Swadesh1952} are available, so the chance of obtaining some concepts missing for individual languages is considerably high. Setting the threshold a bit lower allows us to predefine a core set of concepts that are comparable across languages (and which could be modified anytime, depending on the analysis one desires to conduct). Second, in our modified data aggregation workflow, the threshold is applied to individual language varieties rather than to entire datasets. This means that for all languages in the sample, we count whether they meet the inclusion criterion or not. As a result, it may happen that only certain parts of the datasets from which \clicsfour{} aggregates the word lists make it into the final database. Third, in order to yield a more meaningful selection of concepts, our workflow first orders all concepts by their occurrence across the languages in the data and then retains the most frequent 1,800 concepts. When aggregating the data from the individual word lists, only these concepts are retained. This procedure helps to decrease the sparsity of the data, resulting from the fact that the individual word lists often differ quite drastically with respect to the concepts for which they provide elicitation glosses. {While the cutoff point may seem arbitrary, it reflects our experience in working with the mapping of concept lists in the Concepticon project:  beyond 1,800 concepts, the chances of finding concepts expressed across many languages from many different families drop considerably.}

\subsection{Data Representation}\label{sec:3.4}
The CLICS~3 colexification data was shared in the form of an SQLite database, while the network information was shared in the form of a GML file, offering the colexification networks with nodes, edges, and specific node and edge attributes. It was not a difficult task to implement the CLICS~3 workflow because the GML format can be easily read by different software packages. However, working with the data revealed several shortcomings of the GML format as the exclusive format for sharing the colexification network. 
 
When following the core principle of CLDF in using tables as the basic representation format wherever possible, it would be straightforward to represent a graph with the help of two tables. One table would represent the nodes of a graph, with node attributes being provided in additional columns, and another table would represent the edges, with edge attributes being represented in additional columns. It turned out that this format could not only be easily represented in the CLDF specification, but that it would allow us to represent colexification data in the form of a \emph{structural dataset} \citep{Forkel2018}. While the primary dataset underlying \clicsfour{} provides information on colexifications between a fixed set of standardized concept sets, the additional view as a structural dataset -- resembling a cross-linguistic typological database -- offers a language-centered view: colexifications are modeled as parameters and for each language we provide information on their presence or absence. 
Thus, following \cite{Forkel2020} in combining a word list and a structural dataset in a unified CLDF dataset, \clicsfour{} now consists of a large aggregated word list with individual word forms across several thousand language varieties, along with structural data that provides information on the languages that exhibit certain colexifications.

Structural data in CLDF typically consist of a \emph{parameter table} that provides information on the features comparable across languages, and a \emph{value table} that provides information on the individual values as they are reflected in individual languages. In our new data model for cross-linguistic colexification data, all individual colexifications that can be inferred when analyzing the aggregated word list feature are represented as \emph{parameters}. In contrast, the corresponding values for each language are represented by three different codes, indicating if the feature represented by the parameter is \emph{present}, \emph{absent}, or \emph{missing}. Thus, our proposal for \clicsfour{} not only informs whether a given language exhibits a particular colexification but also whether it does \emph{not} show the colexification, or whether the information is missing, since elicitation glosses for at least one of the concepts involved in the colexification are missing in the word list. 

There are two major advantages of this new representation. The first advantage is that colexifications can be directly inspected in tabular form. Since the colexification data are shared in a table format as part of the CLDF dataset underlying \clicsfour{}, interested users can browse through the colexifications using their favorite spreadsheet editor. Analyzing the colexification network with software tools is also facilitated, given that all major tools support tabular data. This means that networks can be conveniently analyzed computationally or visualized with graph visualization software, such as Cytoscape \citep[for a tutorial, see][]{Tjuka2024TBLOG02}. The second advantage is that it is much easier to integrate the data produced by \clicsfour{} with the data shared by other projects. Community assignments, along with additional information on the coverage of concepts across languages and language families, for example, are now part of the concept table that serves as the basic parameter table for the \clicsfour{} word list. From this representation, it is easy to integrate the data not only into the Concepticon \citep[see also][]{Bocklage2024} but also into extended reference catalogs such as NoRaRe \citep[\href{https://norare.clld.org}{https://norare.clld.org}, ][]{Tjuka2022}, a catalog that extends the Concepticon by providing additional information on \emph{norms}, \emph{rates}, and \emph{ratings} for words and concepts across multiple languages.

\subsection{Implementation}
\clicsfour{} is implemented in the form of a \texttt{CLDFBench} package \citep{Forkel2020}, written in Python, that can be installed from the command line and contains the resulting CLDF data along with the code that was used to create the data. The package is shared as part of the supplemental material accompanying this study and contains additional information and code examples that were used to produce the findings presented in this study. 

\begin{figure*}[tb]
\centering
\includegraphics[width=\textwidth]{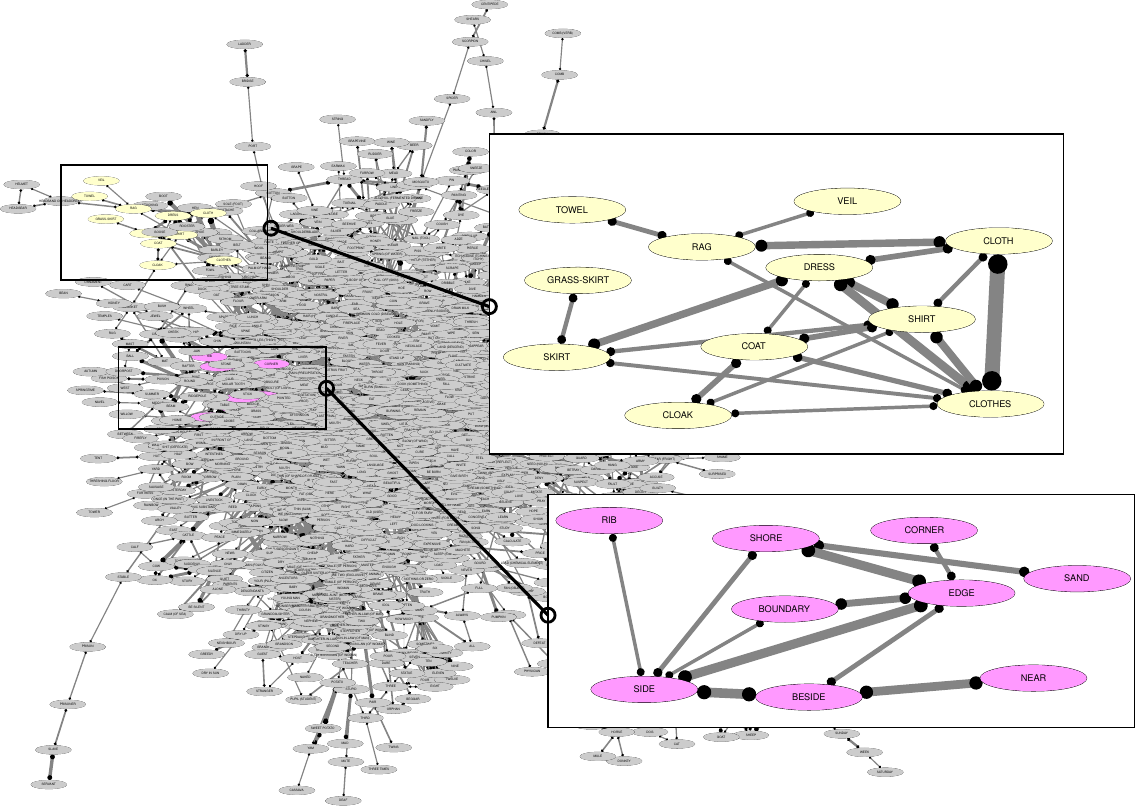}
\caption{\clicsfour{} colexification network with two selected communities (central concepts DRESS and EDGE). }
\label{fig:clicsnetwork}
\end{figure*}

\section{Data Validation}\label{sec:4}

\subsection{Comparing CLICS~3 and \clicsfour{}}
In order to understand the differences between our updated version \clicsfour{} and the previous versions of CLICS, most importantly the last officially published version CLICS~3 by \citet{Rzymski2020}, we carried out
a detailed comparison of CLICS~3 and \clicsfour{}. Given that we deliberately restricted the number of concepts in \clicsfour{} to an initial list of 1,800 concepts -- of which 1,730 were retained when selecting those languages that would cover at least 180 concepts of the initial list -- it may seem as if \clicsfour{} simply reduced the amount of data in contrast to CLICS~3. However, this is not the case, which is apparent when comparing the number of words, language varieties, languages (different glottocodes), and language families covered in both datasets, as shown in Table \ref{tab:stats}. \clicsfour{} exceeds CLICS~3 not only regarding the number of language families and language varieties covered, but most notably with respect to the number of word forms that are provided in phonetic transcriptions. \clicsfour{} reaches almost the same size as CLICS~3, while providing almost three times as many phonetic transcriptions.

A similar situation arises when comparing the overall number of concepts with the average number of languages and families \emph{expressing} a concept in both datasets (also shown in Table \ref{tab:stats}). Here, CLICS~3 exceeds \clicsfour{} in the number of concepts that are colexified (1,386 vs. 1,647), while showing similar values for the average number of languages expressing a concept (607 vs. 624). However, regarding the average number of families expressing a concept, \clicsfour{} largely exceeds CLICS~3 (92 vs. 61). 

In sum, the comparison provided in Table \ref{tab:stats} shows that \clicsfour{} does not simply provide \emph{more} data, resulting in more languages, more concepts, and more colexifications. Instead, the major improvements concerning the data basis, concept handling, and language selection yield a colexification network that consolidates the tendencies in the data rather than diversifying them further. Thus, while \clicsfour{} has fewer colexified concepts, {i.e., concepts that are part of a colexification}, the concepts in the colexification network of \clicsfour{} have more connections across more language families on average, as reflected in their degree distribution (6 vs. 5). In addition, these connections are also substantiated by more {colexifications}, as reflected in the weighted degree distribution (53 vs. 36). 
This trend can also be observed when directly comparing the inferred colexifications. There are 2,874 colexifications observed in both networks, 1,354 unique to CLICS~3, and 1,112 unique to \clicsfour{}. Of the common edges, 859 colexifications in \clicsfour{} can be found in more language families, compared to 778 colexifications in CLICS~3.

\begin{table}
\centering
\resizebox{\linewidth}{!}{%
\begin{tabular}{|l|c|c|}
        \hline
   \textbf{Criterion} & \textbf{CLICS~3}  & \textbf{\clicsfour{}} \\\hline \hline
Datasets & 30 & 95 \\\hline
Varieties & 3~156  & 3~432 \\\hline 
Languages & 2~280& 2~152 \\\hline
Families & 200 & 247 \\\hline
Words & 1~462~125  &  1~445~845 \\\hline 
Transcriptions &  563~878 & 1~445~845 \\\hline 
Words per Variety & 467 &  421 \\\hline \hline
Concepts & 2~906  & 1~730 \\\hline 
Colexified Concepts & 1~647& 1~386 \\\hline
Languages per Concept & 624 & 607 \\\hline 
Families per Concept & 61 & 92 \\\hline \hline
Colexifications & 4~228 & 3~986 \\\hline 
Average Degree & 5& 6\\\hline
Average Weighted Degree & 36 & 53 \\\hline\hline 
Communities  &  249 &  315 \\\hline
Concepts per Community &  6.6 & 4.4 \\  \hline
\end{tabular}}
\caption{Comparison between CLICS~3 and \clicsfour{}. Colexifications are only counted when occurring in at least three different language families. Weighted degree is calculated by counting the number of language families per link.}
\label{tab:stats}
\end{table}


\subsection{Visualizing the \clicsfour{} Network}

To create a visual representation of the \clicsfour{} network, researchers can either use the GML file that is provided along with the CLDF data, or the table with all colexifications that is shared as part of CLDF directly. As mentioned in §~\ref{sec:3.4}, the new data representation in tabular form as part of a unified CLDF dataset makes it easier to analyze the data computationally. The visualization of the data is also greatly facilitated, given that edge tables are the basic input format for network visualization software tools like Cytoscape. A tutorial on how to create a network visualization with Cytoscape is provided in \citealt{Tjuka2024TBLOG02}. We used this approach to create Figure \ref{fig:clicsnetwork}, which provides a bird's eye view of the \clicsfour{} network.  

The figure shows the entire network with two communities highlighted and enlarged. The first community has the concept \href{https://concepticon.clld.org/parameters/474}{\textsc{dress}} as a central node and shows colexifications with other clothing items. The edge weights represent the frequency with which a given colexification occurs across languages. For example, the colexification between \textsc{dress} and \textsc{skirt} is more frequent than the colexification with \textsc{coat}. The second community has the concept \textsc{edge} as a central node and includes cross-linguistically frequent colexifications such as \textsc{edge} and \textsc{side} and less frequent ones like \textsc{edge} and \textsc{corner}. Given the straightforward representation of the colexification network in \clicsfour{}, the data can conveniently be explored. By using Cytoscape, researchers can further investigate the properties of the network and filter them according to their particular research interests.

\section{Conclusion and Outlook}\label{sec:5}

We presented \clicsfour{}, an enhanced version of the Cross-Linguistic Colexification Database, which integrates lexical data for 3,432 language varieties, corresponding to 2,152 distinct Glottocodes. When creating \clicsfour{} we used an advanced workflow for the aggregation and analysis of cross-linguistic colexification data that is based on an increased and improved data basis, an improved handling of concepts,
more fine-grained criteria for the selection of languages and concepts, and an updated representation of the colexification data. 
 
In contrast to previous colexification databases, \clicsfour{} determines colexifications exclusively based on phonetic transcriptions. This makes the data more consistent and robust and opens new possibilities to analyze the data in comparative studies. Due to the phonetic transcriptions in \clicsfour{}, future studies can build on the initial work to infer and investigate partial colexifications \citep{List2023a,Tjuka2024a,Rubehn2025a}. In addition, phonetic transcriptions enable scholars to carry out more fine-grained analyses of colexifications inside specific language families, where a handling of cognate words is important to identify colexifications that have evolved independently from colexifications that have been inherited across branches \citep{Tjuka2024}.


Future studies can use \clicsfour{} to explore the relationship between words and their meanings across a wide range of languages and uncover important insights into language evolution, cultural variations, and cognitive principles. In this way, \clicsfour{} has great potential to contribute to future studies that address open questions in a broad range of linguistic subfields, including linguistic typology, historical linguistics, psycholinguistics, and computational linguistics. 

\section*{Supplementary Material} 

All data and code underlying this study, along with instructions on how to run the code, are openly available. The \clicsfour{} database is curated on GitHub (\href{https://github.com/clics/clics4/tree/v0.5}{https://github.com/clics/clics4/tree/v0.5}, Version 0.5) and archived with Zenodo (DOI: \href{https://doi.org/10.5281/zenodo.16900180}{https://doi.org/10.5281/zenodo.16900180}). The code that we used to compare CLICS 3 and \clicsfour{} is cuarated on Codeberg (\href{https://codeberg.org/calc/clics4-paper/src/tag/v1.0}{https://codeberg.org/calc/clics4-paper/src/tag/v1.0}, Version 1.0) and archived with Zenodo (DOI: \href{https://doi.org/10.5281/zenodo.16902185}{https://doi.org/10.5281/zenodo.16902185}).

\section*{Limitations}
General limitations that apply to large-scale aggregation studies in comparative linguistics also apply to \clicsfour{}. These include the fact that the word list approach for aggregating cross-linguistic colexifications may fail to model fine-grained aspects of colexifications in individual language families, many of which cannot be modeled appropriately without a detailed inspection of particular languages and their history. An additional problem of all cross-linguistic colexification databases is that they contain a lot of missing data, showing low coverage for most concepts cross-linguistically. We also emphasize that detailed studies investigating the properties of \clicsfour{} are missing so far, but we envisage that these will be carried out by different teams (not only including the team which compiled the data by now). Another improvement that needs to be implemented in the future is the treatment of some artificially separated concepts. For example, the current version splits the concept \href{https://concepticon.clld.org/parameters/2271}{\textsc{think}} into the more specific concepts \href{https://concepticon.clld.org/parameters/1415}{\textsc{think (reflect)}} and \href{https://concepticon.clld.org/parameters/1513}{\textsc{think (believe)}}. While this modification reflects the ambiguity of the concept \textsc{think}, we suspect that there is no frequently used questionnaire for cross-linguistic data that would contain both \textsc{think (reflect)} and \textsc{think (believe)}. As a result, one may call the colexification between \textsc{think (reflect)} and \textsc{think (believe)} in question, given that the database lacks direct evidence. This holds to an even larger degree for kinship terms.
One solution we could think of would be to consider only \textsc{think}, as the broadest concept, because this concept is present in most languages. While our current technology would allow for such a handling, we think addressing this problem in a principled way will require a more thorough revision, potentially accompanied by additional computational analyses and very detailed decisions that should not be made in an ad-hoc style.

So far, \clicsfour{} is limited to the data and the database itself can only be investigated with tools for network visualization and with computational approaches. 
As of now, the web application at \href{https://clics.clld.org}{https://clics.clld.org} still serves the data underlying CLICS~3. Implementing the web application for \clicsfour{} is planned and will follow in the near future.

\section*{Acknowledgments}

This project was supported by the ERC Consolidator Grant ProduSemy (JML; Grant No. 101044282, see \href{https://doi.org/10.3030/101044282}{https://doi.org/10.3030/101044282}) and the ERC Synergy Grant QUANTA (CR; Grant No. 951388, see \href{https://doi.org/10.3030/951388}{https://doi.org/10.3030/951388}). Views and opinions expressed are, however, those of the author(s) only and do not necessarily reflect those of the European Union or the European Research Council Executive Agency (nor any other funding agencies involved). Neither the European Union nor the granting authority can be held responsible for them.




\bibliography{custom}

\begin{thebibliography}{59}
\providecommand{\natexlab}[1]{#1}

\bibitem[{Anderson et~al.(2018)Anderson, Tresoldi, Chacon, Fehn, Walworth, Forkel, and List}]{Anderson2018}
Cormac Anderson, Tiago Tresoldi, Thiago~Costa Chacon, Anne-Maria Fehn, Mary Walworth, Robert Forkel, and Johann-Mattis List. 2018.
\newblock \href {https://doi.org/10.2478/yplm-2018-0002} {{A} cross-linguistic database of phonetic transcription systems}.
\newblock \emph{Yearbook of the Poznań Linguistic Meeting}, 4(1):21--53.

\bibitem[{Apresjan(1974)}]{Apresjan1974}
Juri~D. Apresjan. 1974.
\newblock \href {https://doi.org/10.1515/ling.1974.12.142.5} {Regular polysemy}.
\newblock \emph{Linguistics}, 12(142):5--32.

\bibitem[{Bao et~al.(2021)Bao, Hauer, and Kondrak}]{Bao2021}
Hongchang Bao, Bradley Hauer, and Grzegorz Kondrak. 2021.
\newblock \href {https://aclanthology.org/2021.gwc-1.1.pdf} {On universal colexifications}.
\newblock In \emph{Proceedings of the 11th {{Global WordNet Conference}}}, pages 1--7, University of South Africa. Global WordNet Association.

\bibitem[{Bao et~al.(2022)Bao, Hauer, and Kondrak}]{Bao2022}
Hongchang Bao, Bradley Hauer, and Grzegorz Kondrak. 2022.
\newblock \href {https://aclanthology.org/2022.lrec-1.774/} {Lexical resource mapping via translations}.
\newblock In \emph{Proceedings of the Thirteenth Language Resources and Evaluation Conference}, pages 7147--7154, Marseille, France. European Language Resources Association.

\bibitem[{Blevins and Sproat(2021)}]{Blevins2021}
Juliette Blevins and Richard Sproat. 2021.
\newblock \href {https://doi.org/10.1075/dia.19014.ble} {Statistical evidence for the {Proto-Indo-European-Euskarian Hypothesis}: {A} word-list approach integrating phonotactics}.
\newblock \emph{Diachronica}, 0(0):1--59.

\bibitem[{Blum et~al.(2025)Blum, Barrientos, Englisch, Forkel, Greenhill, Rzymski, and List}]{Blum2025b}
Frederic Blum, Carlos Barrientos, Johannes Englisch, Robert Forkel, Simon~J. Greenhill, Christoph Rzymski, and Johann-Mattis List. 2025.
\newblock \href {https://doi.org/10.12688/openreseurope.20216.2} {Lexibank 2: Pre-computed features for large-scale lexical data {[version 2; peer review: 3 approved]}}.
\newblock \emph{Open Research Europe}, 5(126):1--24.

\bibitem[{Blum et~al.(2024)Blum, Barrientos, Ingunza, and List}]{Blum2024d}
Frederic Blum, Carlos Barrientos, Adriano Ingunza, and Johann-Mattis List. 2024.
\newblock \href {https://doi.org/10.1038/s41598-024-82515-3} {Cognate reflex prediction as hypothesis test for a genealogical relation between the {Panoan and Takanan} language families}.
\newblock \emph{Scientific Reports}, 14(30636):1--12.

\bibitem[{Bocklage et~al.(2024)Bocklage, Di~Natale, Tjuka, and List}]{Bocklage2024}
Katja Bocklage, Anna Di~Natale, Annika Tjuka, and Johann-Mattis List. 2024.
\newblock \href {https://doi.org/10.17613/0y0r-f341} {\emph{Directional tendencies in semantic change}}.
\newblock Humanities Commons.

\bibitem[{Bodt and List(2019)}]{Bodt2019}
Timotheus~A. Bodt and Johann-Mattis List. 2019.
\newblock \href {https://doi.org/10.2218/pihph.4.2019.3037} {Testing the predictive strength of the comparative method: {A}n ongoing experiment on unattested words in {{Western Kho}}-{{Bwa languages}}}.
\newblock \emph{Papers in Historical Phonology}, 4:22--44.

\bibitem[{Bowern and Atkinson(2012)}]{Bowern2012}
Claire Bowern and Quentin Atkinson. 2012.
\newblock \href {https://doi.org/10.1353/lan.2012.0081} {Computational {{phylogenetics}} and the {{internal structure}} of {{Pama-Nyungan}}}.
\newblock \emph{Language}, 88(4):817--845.

\bibitem[{Bradford et~al.(2022)Bradford, Thomas, and Xu}]{Bradford2022}
Laurestine Bradford, Guillaume Thomas, and Yang Xu. 2022.
\newblock \href {https://escholarship.org/uc/item/7x681267} {Communicative need modulates lexical precision across semantic domains: {A} domain-level account of efficient communication}.
\newblock In \emph{Proceedings of the Annual Meeting of the Cognitive Science Society}, pages 2561--2568.

\bibitem[{Brochhagen and Boleda(2022)}]{Brochhagen2022}
Thomas Brochhagen and Gemma Boleda. 2022.
\newblock \href {https://doi.org/10.1016/j.cognition.2022.105179} {When do languages use the same word for different meanings? {T}he {Goldilocks Principle} in colexification}.
\newblock \emph{Cognition}, 226:1--8.

\bibitem[{Cs{\'a}rdi and Nepusz(2006)}]{Csardi2006}
G{\'a}bor Cs{\'a}rdi and Tam{\'a}s Nepusz. 2006.
\newblock \href {https://doi.org/10.5281/zenodo.3630268} {The {Igraph} software package for complex network research}.
\newblock \emph{InterJournal Complex Systems}, 1695.

\bibitem[{Cysouw(2010)}]{Cysouw2010}
Michael Cysouw. 2010.
\newblock \href {http://journals.dartmouth.edu/cgi-bin/WebObjects/Journals.woa/1/xmlpage/1/article/377} {{D}rawing networks from recurrent polysemies}.
\newblock \emph{Linguistic Discovery}, 8(1):281--285.

\bibitem[{Di~Natale et~al.(2021)Di~Natale, Pellert, and Garcia}]{DiNatale2021}
Anna Di~Natale, Max Pellert, and David Garcia. 2021.
\newblock \href {https://doi.org/10.1007/s42761-021-00033-1} {Colexification networks encode affective meaning}.
\newblock \emph{Affective Science}, 2:99--111.

\bibitem[{Enfield and Comrie(2015)}]{Enfield2015}
Nick~J. Enfield and Bernard Comrie, editors. 2015.
\newblock \emph{{Languages of Mainland South-East Asia. The state of the art}}.
\newblock Mouton de Gruyter, Berlin and New York.

\bibitem[{Forkel and List(2020)}]{Forkel2020}
Robert Forkel and Johann-Mattis List. 2020.
\newblock \href {http://www.lrec-conf.org/proceedings/lrec2020/pdf/2020.lrec-1.864.pdf} {{{CLDFBench}}: {{Give your cross-linguistic data}} a {{lift}}}.
\newblock In \emph{Proceedings of the 12th {{Language Resources}} and {{Evaluation Conference}}}, pages 6995--7002, Marseille, France. European Language Resources Association.

\bibitem[{Forkel et~al.(2018)Forkel, List, Greenhill, Rzymski, Bank, Cysouw, Hammarstr{\"o}m, Haspelmath, Kaiping, and Gray}]{Forkel2018}
Robert Forkel, Johann-Mattis List, Simon~J. Greenhill, Christoph Rzymski, Sebastian Bank, Michael Cysouw, Harald Hammarstr{\"o}m, Martin Haspelmath, Gereon~A. Kaiping, and Russell~D. Gray. 2018.
\newblock \href {https://doi.org/10.1038/sdata.2018.205} {Cross-{{Linguistic Data Formats}}, advancing data sharing and re-use in comparative linguistics}.
\newblock \emph{Scientific Data}, 5(1):1--10.

\bibitem[{Fran{\c c}ois(2008)}]{Francois2008}
Alexandre Fran{\c c}ois. 2008.
\newblock \href {https://doi.org/10.1075/slcs.106.09fra} {Semantic maps and the typology of colexification: {{Intertwining polysemous networks across languages}}}.
\newblock In Martine Vanhove, editor, \emph{From {{polysemy}} to {{semantic change}}: {{Towards}} a typology of lexical semantic associations}, pages 163--215. John Benjamins, Amsterdam.

\bibitem[{Gast and {Koptjevskaja-Tamm}(2019)}]{Gast2019}
Volker Gast and Maria {Koptjevskaja-Tamm}. 2019.
\newblock \href {https://doi.org/10.1515/9783110607963-003} {The areal factor in lexical typology: {{Some evidence}} from {{lexical databases}}}.
\newblock In Dani{\"e}l Van~Olmen, Tanja Mortelmans, and Frank Brisard, editors, \emph{Aspects of {{Linguistic Variation}}}, pages 43--82. Walter de Gruyter, Berlin.

\bibitem[{Gower(2021)}]{CSVW}
Robin Gower. 2021.
\newblock \href {https://csvw.org} {\emph{{CSV} on the Web}}.
\newblock Swirrl, Stirling.

\bibitem[{Hagberg et~al.(2008)Hagberg, Schult, and Swart}]{Hagberg2008}
Aric~A. Hagberg, Daniel~A. Schult, and Pieter~J. Swart. 2008.
\newblock \href {https://networkx.org} {Exploring network structure, dynamics, and function using {NetworkX}}.
\newblock In \emph{Proceedings of the 7th {{Python}} in {{Science Conference}}}, pages 11--15, Pasadena.

\bibitem[{Hammarström et~al.(2025)Hammarström, Haspelmath, Forkel, and Bank}]{Glottolog}
Harald Hammarström, Martin Haspelmath, Robert Forkel, and Sebastian Bank. 2025.
\newblock \href {https://glottolog.org} {\emph{{G}lottolog [{Dataset, Version 5.2.1}]}}.
\newblock Max Planck Institute for Evolutionary Anthropology, Leipzig.

\bibitem[{Haspelmath and Tadmor(2009)}]{Haspelmath2009}
Martin Haspelmath and Uri Tadmor. 2009.
\newblock The {{Loanword Typology Project}} and the {{World Loanword Database}}.
\newblock In Martin Haspelmath and Uri Tadmor, editors, \emph{Loanwords in the {{World}}'s {{Languages}}}, pages 1--34. De Gruyter Mouton, Berlin.

\bibitem[{Himsolt(2010)}]{Himsolt2010}
Michael Himsolt. 2010.
\newblock \href {http://www.fim.uni-passau.de/fileadmin/files/lehrstuhl/brandenburg/projekte/gml/gml-technical-report.pdf} {{GML}: {A} portable graph file format}.
\newblock Technical report, Universität Passau.

\bibitem[{{IPA, International Phonetic Association}(1999)}]{IPA1999}
{IPA, International Phonetic Association}. 1999.
\newblock \emph{{H}andbook of the {I}nternational {P}honetic {A}ssociation}.
\newblock Cambridge University Press, Cambridge.

\bibitem[{Jackson et~al.(2019)Jackson, Watts, Henry, List, Forkel, Mucha, Greenhill, Gray, and Lindquist}]{Jackson2019}
Joshua~Conrad Jackson, Joseph Watts, Teague~R. Henry, Johann-Mattis List, Robert Forkel, Peter~J. Mucha, Simon~J. Greenhill, Russell~D. Gray, and Kristen~A. Lindquist. 2019.
\newblock \href {https://doi.org/10.1126/science.aaw8160} {Emotion semantics show both cultural variation and universal structure}.
\newblock \emph{Science}, 366:1517--1522.

\bibitem[{Key and Comrie(2023)}]{Key2023}
Mary~Ritchie Key and Bernard Comrie. 2023.
\newblock \href {https://ids.clld.org} {\emph{The {{Intercontinental Dictionary Series}} [{{Dataset, Version}} 4.3]}}.
\newblock Max Planck Institute for Evolutionary Anthropology, Leipzig.

\bibitem[{Leivada and Murphy(2021)}]{Leivada2021}
Evelina Leivada and Elliot Murphy. 2021.
\newblock \href {https://doi.org/10.1016/j.amper.2021.100073} {Mind the (terminological) gap: 10 misused, ambiguous, or polysemous terms in linguistics}.
\newblock \emph{Ampersand}, 8:1--9.

\bibitem[{List(2022)}]{List2022TBLOG06}
Johann-Mattis List. 2022.
\newblock \href {https://calc.hypotheses.org/4266} {How to compute colexifications with {CL Toolkit} ({H}ow to do {X} in linguistics 10)}.
\newblock \emph{Computer-Assisted Language Comparison in Practice}, 5(6).

\bibitem[{List(2023)}]{List2023a}
Johann-Mattis List. 2023.
\newblock \href {https://doi.org/10.3389/fpsyg.2023.1156540} {Inference of partial colexifications from multilingual wordlists}.
\newblock \emph{Frontiers in Psychology}, 14:1--10.

\bibitem[{List et~al.(2021)List, Anderson, Tresoldi, and Forkel}]{CLTS}
Johann-Mattis List, Cormac Anderson, Tiago Tresoldi, and Robert Forkel. 2021.
\newblock \href {https://clts.clld.org} {\emph{{C}ross-{L}inguistic {T}ranscription {S}ystems [{Dataset, Version 2.3.0]}}}.
\newblock Max Planck Institute for the Science of Human History, Jena.

\bibitem[{List et~al.(2016)List, Cysouw, and Forkel}]{List2016}
Johann-Mattis List, Michael Cysouw, and Robert Forkel. 2016.
\newblock \href {https://aclanthology.org/L16-1379/} {Concepticon: {{A resource}} for the {{linking}} of {{concept lists}}}.
\newblock In \emph{Proceedings of the 10th {{International Conference}} on {{Language Resources}} and {{Evaluation}}}, pages 2393--2400, Portoro{\v z}, Slovenia. European Language Resources Association.

\bibitem[{List et~al.(2022)List, Forkel, Greenhill, Rzymski, Englisch, and Gray}]{List2022}
Johann-Mattis List, Robert Forkel, Simon~J. Greenhill, Christoph Rzymski, Johannes Englisch, and Russell~D. Gray. 2022.
\newblock \href {https://doi.org/10.1038/s41597-022-01432-0} {Lexibank, a public repository of standardized wordlists with computed phonological and lexical features}.
\newblock \emph{Scientific Data}, 9(1):1--16.

\bibitem[{List et~al.(2018)List, Greenhill, Anderson, Mayer, Tresoldi, and Forkel}]{List2018a}
Johann-Mattis List, Simon~J. Greenhill, Cormac Anderson, Thomas Mayer, Tiago Tresoldi, and Robert Forkel. 2018.
\newblock \href {https://doi.org/10.1515/lingty-2018-0010} {{{CLICS}}{$^2$}: An improved database of cross-linguistic colexifications assembling lexical data with the {{help}} of {{Cross-Linguistic Data Formats}}}.
\newblock \emph{Linguistic Typology}, 22(2):277--306.

\bibitem[{List et~al.(2014)List, Mayer, Terhalle, and Urban}]{List2014}
Johann-Mattis List, Thomas Mayer, Anselm Terhalle, and Matthias Urban. 2014.
\newblock \href {https://clics.lingpy.org} {\emph{{CLICS}: {D}atabase of {C}ross-{L}inguistic {C}olexifications {[Dataset, Version 1.0]}}}.
\newblock Forschungszentrum Deutscher Sprachatlas, Marburg.

\bibitem[{List et~al.(2013)List, Terhalle, and Urban}]{List2013}
Johann-Mattis List, Anselm Terhalle, and Matthias Urban. 2013.
\newblock \href {https://aclanthology.org/W13-0208} {Using network approaches to enhance the analysis of cross-linguistic polysemies}.
\newblock In \emph{Proceedings of the 10th {{International Conference}} on {{Computational Semantics}}}, pages 347--353, Potsdam, Germany. Association for Computational Linguistics.

\bibitem[{List et~al.(2025)List, Tjuka, Blum, Kučerová, Barrientos~Ugarte, Rzymski, Greenhill, and Forkel}]{Concepticon}
Johann-Mattis List, Annika Tjuka, Frederic Blum, Alžběta Kučerová, Carlos Barrientos~Ugarte, Christoph Rzymski, Simon~J. Greenhill, and Robert Forkel. 2025.
\newblock \href {https://concepticon.clld.org/} {\emph{{CLLD Concepticon} [{Dataset, Version 3.4.0}]}}.
\newblock Max Planck Institute for Evolutionary Anthropology, Leipzig.

\bibitem[{Mayer et~al.(2014)Mayer, List, Terhalle, and Urban}]{Mayer2014}
Thomas Mayer, Johann-Mattis List, Anselm Terhalle, and Matthias Urban. 2014.
\newblock \href {https://lingulist.de/documents/mayer-et-al-2014-clics-visualization.pdf} {An interactive visualization of crosslinguistic colexification patterns}.
\newblock In \emph{Proceedings of the {{LREC Workshop}} '{{VisLR}}: {{Visualization}} as {{Added Value}} in the {{Development}}, {{Use}} and {{Evaluation}} of {{Language Resources}}'}, pages 1--8, Reykjavik, Iceland. European Language Resources Association.

\bibitem[{Miller and List(2024)}]{Miller2024TBLOG09}
John Miller and Johann-Mattis List. 2024.
\newblock \href {https://doi.org/10.15475/calcip.2024.2.3} {Adding standardized transcriptions to {Panoan and Tacanan} languages in the {Intercontinental Dictionary Series}}.
\newblock \emph{Computer-Assisted Language Comparison in Practice}, 7(2):69–77.

\bibitem[{Miller et~al.(2020)Miller, Tresoldi, Zariquiey, Beltrán~Castañón, Morozova, and List}]{Miller2020}
John~E. Miller, Tiago Tresoldi, Roberto Zariquiey, César~A. Beltrán~Castañón, Natalia Morozova, and Johann-Mattis List. 2020.
\newblock \href {https://doi.org/10.1371/journal.pone.0242709} {Using lexical language models to detect borrowings in monolingual wordlists}.
\newblock \emph{PLOS One}, 15(12):e0242709.

\bibitem[{Newman(2006)}]{Newman2006b}
M.~E.~J. Newman. 2006.
\newblock \href {https://doi.org/10.1073/pnas.0601602103} {Modularity and community structure in networks}.
\newblock \emph{Proceedings of the National Academy of Science of the United States of America}, 103(23):8577--8582.

\bibitem[{Rosvall and Bergstrom(2008)}]{Rosvall2008}
Martin Rosvall and Carl~T. Bergstrom. 2008.
\newblock \href {https://doi.org/10.1073/pnas.0706851105} {Maps of random walks on complex networks reveal community structure}.
\newblock \emph{Proceedings of the National Academy of Sciences}, 105(4):1118--1123.

\bibitem[{Rubehn and List(2025)}]{Rubehn2025a}
Arne Rubehn and Johann-Mattis List. 2025.
\newblock \href {https://aclanthology.org/2025.acl-long.1004} {Partial colexifications improve concept embeddings}.
\newblock In \emph{Proceedings of the Association for Computational Linguistics 2025. Long Papers}, pages 20571--20586.

\bibitem[{Rubehn et~al.(2024)Rubehn, Nieder, Forkel, and List}]{Rubehn2024a}
Arne Rubehn, Jessica Nieder, Robert Forkel, and Johann-Mattis List. 2024.
\newblock \href {https://doi.org/10.7275/scil.2144} {Generating feature vectors from phonetic transcriptions in {Cross-Linguistic Data Formats}}.
\newblock \emph{Proceedings of the Society for Computation in Linguistics}, 7(1):205--216.

\bibitem[{Rzymski et~al.(2020)Rzymski, Tresoldi, Greenhill, Wu, Schweikhard, {Koptjevskaja-Tamm}, Gast, Bodt, Hantgan, Kaiping, Chang, Lai, Morozova, Arjava, H{\"u}bler, Koile, Pepper, Proos, Van~Epps, Blanco, Hundt, Monakhov, Pianykh, Ramesh, Gray, Forkel, and List}]{Rzymski2020}
Christoph Rzymski, Tiago Tresoldi, Simon~J. Greenhill, Mei-Shin Wu, Nathanael Schweikhard, Maria {Koptjevskaja-Tamm}, Volker Gast, Timotheus~A. Bodt, Abbie Hantgan, Gereon~A. Kaiping, Sophie Chang, Yunfan Lai, Natalia Morozova, Heini Arjava, Nataliia H{\"u}bler, Ezequiel Koile, Steve Pepper, Mariann Proos, Briana Van~Epps, Ingrid Blanco, Carolin Hundt, Sergei Monakhov, Kristina Pianykh, Sallona Ramesh, Russell~D. Gray, Robert Forkel, and Johann-Mattis List. 2020.
\newblock \href {https://doi.org/10.1038/s41597-019-0341-x} {The {{Database}} of {{Cross-Linguistic Colexifications}}, reproducible analysis of cross-linguistic polysemies}.
\newblock \emph{Scientific Data}, 7(1):1--12.

\bibitem[{Schapper(2019)}]{Schapper2019}
Antoinette Schapper. 2019.
\newblock \href {https://doi.org/10.1353/ol.2019.0004} {The ethno-linguistic relationship between smelling and kissing: {A} {Southeast Asian} case study}.
\newblock \emph{Oceanic Linguistics}, 58(1):92--109.

\bibitem[{Schapper(2022)}]{Schapper2022a}
Antoinette Schapper. 2022.
\newblock \href {https://doi.org/10.1515/lingty-2021-2082} {Baring the bones: {{The}} lexico-semantic association of bone with strength in {{Melanesia}} and the study of {{colexification}}}.
\newblock \emph{Linguistic Typology}, 26(2):313--347.

\bibitem[{Sjöberg(2023)}]{Sjoeberg2023}
Anna Sjöberg. 2023.
\newblock \href {https://su.diva-portal.org/smash/get/diva2:1800727/FULLTEXT02.pdf} {\emph{Knowledge predication: A semantic typology}}.
\newblock Ph.D. thesis, Stockholm University, Stockholm.

\bibitem[{Smoot et~al.(2011)Smoot, Ono, Ruscheinski, Wang, and Ideker}]{Smoot2011}
Michael~E. Smoot, Keiichiro Ono, Johannes Ruscheinski, Peng-Liang Wang, and Trey Ideker. 2011.
\newblock \href {https://doi.org/10.1093/bioinformatics/btq675} {Cytoscape 2.8: {N}ew features for data integration and network visualization}.
\newblock \emph{Bioinformatics}, 27(3):431--432.

\bibitem[{Souag(2022)}]{Souag2022}
Lameen Souag. 2022.
\newblock \href {https://doi.org/10.1515/lingty-2021-2083} {How a {West African} language becomes {North African}, and vice versa}.
\newblock \emph{Linguistic Typology}, 26(2):283--312.

\bibitem[{Sperber(1923)}]{Sperber1923}
Hans Sperber. 1923.
\newblock \emph{{Einführung in die Bedeutungslehre [Introduction to the study of meaning]}}.
\newblock Kurt Schroeder, Bonn and Leipzig.

\bibitem[{Swadesh(1952)}]{Swadesh1952}
Morris Swadesh. 1952.
\newblock {L}exico-statistic dating of prehistoric ethnic contacts.
\newblock \emph{Proceedings of the American Philosophical Society}, 96(4):452--463.

\bibitem[{Tjuka(2024{\natexlab{a}})}]{Tjuka2024TBLOG02}
Annika Tjuka. 2024{\natexlab{a}}.
\newblock \href {https://doi.org/10.15475/calcip.2024.1.2} {How to visualize colexification networks in {Cytoscape} ({H}ow to do {X} in linguistics 14)}.
\newblock \emph{Computer-Assisted Language Comparison in Practice}, 7(1):7–16.

\bibitem[{Tjuka(2024{\natexlab{b}})}]{Tjuka2024b}
Annika Tjuka. 2024{\natexlab{b}}.
\newblock \href {https://doi.org/10.1515/lingty-2023-0032} {Objects as human bodies: Cross-linguistic colexifications between words for body parts and {{objects}}}.
\newblock \emph{Linguistic Typology}, pages 1--40.

\bibitem[{Tjuka et~al.(2022)Tjuka, Forkel, and List}]{Tjuka2022}
Annika Tjuka, Robert Forkel, and Johann-Mattis List. 2022.
\newblock \href {https://doi.org/10.3758/s13428-021-01650-1} {Linking norms, ratings, and relations of words and concepts across multiple language varieties}.
\newblock \emph{Behavior Research Methods}, 54(2):864–884.

\bibitem[{Tjuka et~al.(2023)Tjuka, Forkel, and List}]{Tjuka2023}
Annika Tjuka, Robert Forkel, and Johann-Mattis List. 2023.
\newblock \href {https://doi.org/10.12688/openreseurope.15380.3} {Curating and {{extending data}} for language comparison in {{Concepticon}} and {{NoRaRe}}}.
\newblock \emph{Open Research Europe}, 2(141):1--13.

\bibitem[{Tjuka et~al.(2024)Tjuka, Forkel, and List}]{Tjuka2024}
Annika Tjuka, Robert Forkel, and Johann-Mattis List. 2024.
\newblock \href {https://doi.org/10.1038/s41598-024-61140-0} {Universal and cultural factors shape body part vocabularies}.
\newblock \emph{Scientific Reports}, 14(1):1--12.

\bibitem[{Tjuka and List(2024)}]{Tjuka2024a}
Annika Tjuka and Johann-Mattis List. 2024.
\newblock \href {https://doi.org/10.1515/gcla-2024-0005} {Partial colexifications reveal directional tendencies in object naming}.
\newblock \emph{Yearbook of the German Cognitive Linguistics Association}, 12(1):95--114.

\end{thebibliography}

\onecolumn
\appendix

\section{Original and Replaced Concepts}
\label{sec:replacements}

\begin{longtable}{|p{4cm}|p{1.75cm}|p{1.75cm}|p{7cm}|}
\hline
 \bfseries Original Concept                 &   \bfseries Rep. Con. &   \bfseries New Con.& \bfseries Details                                                                                                                        \\\hline\hline
\hline
 MOUNTAIN OR HILL                 &           1466 &              2 & HILL (733), MOUNTAIN (733)                                                                                                     \\\hline
 SPRING OR WELL                   &           1016 &              2 & SPRING (OF WATER) (508), WELL (508)                                                                                            \\\hline
 STONE OR ROCK                    &            484 &              1 & STONE (484)                                                                                                                    \\\hline
 MAN                              &           2280 &              1 & MALE PERSON (2280)                                                                                                             \\\hline
 BROTHER                          &           2668 &              4 & OLDER BROTHER (OF MAN) (667), OLDER BROTHER (OF WOMAN) (667), YOUNGER BROTHER (OF MAN) (667), YOUNGER BROTHER (OF WOMAN) (667) \\\hline
 OLDER BROTHER                    &           2462 &              2 & OLDER BROTHER (OF MAN) (1231), OLDER BROTHER (OF WOMAN) (1231)                                                                 \\\hline
 YOUNGER BROTHER                  &           1814 &              2 & YOUNGER BROTHER (OF MAN) (907), YOUNGER BROTHER (OF WOMAN) (907)                                                               \\\hline
 SISTER                           &           3060 &              4 & OLDER SISTER (OF MAN) (765), OLDER SISTER (OF WOMAN) (765), YOUNGER SISTER (OF MAN) (765), YOUNGER SISTER (OF WOMAN) (765)     \\\hline
 OLDER SISTER                     &           1910 &              2 & OLDER SISTER (OF MAN) (955), OLDER SISTER (OF WOMAN) (955)                                                                     \\\hline
 YOUNGER SISTER                   &           1664 &              2 & YOUNGER SISTER (OF MAN) (832), YOUNGER SISTER (OF WOMAN) (832)                                                                 \\\hline
 UNCLE                            &           1254 &              2 & MATERNAL UNCLE (MOTHER'S BROTHER) (627), PATERNAL UNCLE (FATHER'S BROTHER) (627)                                               \\\hline
 AUNT                             &           1406 &              2 & MATERNAL AUNT (MOTHER'S SISTER) (703), PATERNAL AUNT (FATHER'S SISTER) (703)                                                   \\\hline
 HE OR SHE OR IT                  &           3444 &              3 & HE (1148), IT (1148), SHE (1148)                                                                                               \\\hline
 WE                               &           3862 &              2 & WE (EXCLUSIVE) (1931), WE (INCLUSIVE) (1931)                                                                                   \\\hline
 BLOOD VESSEL                     &            342 &              1 & VEIN (342)                                                                                                                     \\\hline
 ROAST OR FRY                     &            868 &              2 & FRY (434), ROAST (SOMETHING) (434)                                                                                             \\\hline
 SIEVE OR STRAIN                  &            409 &              1 & STRAIN (409)                                                                                                                   \\\hline
 TORCH OR LAMP                    &            400 &              1 & LAMP (400)                                                                                                                     \\\hline
 SICKLE OR SCYTHE                 &            445 &              1 & SICKLE (445)                                                                                                                   \\\hline
 BRANCH OR TWIG                   &            353 &              1 & BRANCH (353)                                                                                                                   \\\hline
 STRIKE OR BEAT                   &           1416 &              2 & BEAT (708), STRIKE (708)                                                                                                       \\\hline
 CHOP                             &           1116 &              2 & CHOP (INTO PIECES) (558), CUT (WITH AXE) (558)                                                                                 \\\hline
 BREAK (DESTROY OR GET DESTROYED) &           2240 &              2 & BREAK (BREAKING) (1120), BREAK (CLEAVE) (1120)                                                                                 \\\hline
 TWIST (AROUND)                   &            415 &              1 & TWIST (415)                                                                                                                    \\\hline
 CRAWL OR CREEP                   &            455 &              1 & CRAWL (455)                                                                                                                    \\\hline
 STORE                            &            311 &              1 & SHOP (311)                                                                                                                     \\\hline
 AFTER                            &            743 &              1 & AFTERWARDS (743)                                                                                                               \\\hline
 OLD                              &           5164 &              2 & OLD (AGED) (2582), OLD (USED) (2582)                                                                                           \\\hline
 BREATH OR BREATHE                &            728 &              2 & BREATH (364), BREATHE (364)                                                                                                    \\\hline
 BE ALIVE OR LIFE                 &            990 &              2 & BE ALIVE (495), LIFE (495)                                                                                                     \\\hline
 BE DEAD OR DIE                   &           1358 &              1 & DIE (1358)                                                                                                                     \\\hline
 MIGHTY OR POWERFUL OR STRONG     &            852 &              2 & POWERFUL (426), STRONG (426)                                                                                                   \\\hline
 COOKING POT                      &            660 &              1 & POT (660)                                                                                                                      \\\hline
 DO OR MAKE                       &           1582 &              2 & DO (791), MAKE (791)                                                                                                           \\\hline
 BRONZE OR COPPER                 &            273 &              1 & COPPER (273)                                                                                                                   \\\hline
 DOWN OR BELOW                    &            646 &              2 & BELOW OR UNDER (323), DOWN (323)                                                                                               \\\hline
 CENTER OR MIDDLE                 &            337 &              1 & MIDDLE (337)                                                                                                                   \\\hline
 BEGIN OR START                   &            520 &              1 & BEGIN (520)                                                                                                                    \\\hline
 CANNON OR GUN                    &            338 &              1 & GUN (338)                                                                                                                      \\\hline
 FINGERNAIL OR TOENAIL            &            872 &              2 & FINGERNAIL (436), TOENAIL (436)                                                                                                \\\hline
 PATH OR ROAD                     &           2920 &              2 & PATH (1460), ROAD (1460)                                                                                                       \\\hline
 COLD (OF WEATHER)                &            204 &              1 & COLD (204)                                                                                                                     \\\hline
 A LITTLE                         &            191 &              1 & FEW (191)                                                                                                                      \\\hline
 HOW MANY                         &           1592 &              2 & HOW MANY PIECES (796), HOW MUCH (796)                                                                                          \\\hline
 SON-IN-LAW                       &            434 &              2 & SON-IN-LAW (OF MAN) (217), SON-IN-LAW (OF WOMAN) (217)                                                                         \\\hline
 CUT (WITH KNIFE)                 &            250 &              1 & CUT (250)                                                                                                                      \\\hline
 MARRY (AS MAN)                   &            269 &              1 & MARRY (269)                                                                                                                    \\\hline
 HIT                              &           2051 &              1 & STRIKE (2051)                                                                                                                  \\\hline
 THIN (OF LEAF AND CLOTH)         &            240 &              1 & THIN (OF SHAPE OF OBJECT) (240)                                                                                                \\\hline
 ITCH OR ITCHY OR ITCHING         &            344 &              2 & ITCH (172), ITCH (CAUSE ITCHING OR FEEL ITCHY) (172)                                                                           \\\hline
 HE OR SHE                        &           2052 &              2 & HE (1026), SHE (1026)                                                                                                          \\\hline
 THIN                             &           3456 &              2 & THIN (OF SHAPE OF OBJECT) (1728), THIN (SLIM) (1728)                                                                           \\\hline
 MALE                             &            938 &              2 & MALE (OF ANIMAL) (469), MALE (OF PERSON) (469)                                                                                 \\\hline
 FEMALE PERSON                    &           1154 &              1 & WOMAN (1154)                                                                                                                   \\\hline
 CHILD                            &           3876 &              2 & CHILD (DESCENDANT) (1938), CHILD (YOUNG HUMAN) (1938)                                                                          \\\hline
 HIDE                             &           2594 &              2 & HIDE (CONCEAL) (1297), HIDE (ONESELF) (1297)                                                                                   \\\hline
 THINK                            &           3834 &              2 & THINK (BELIEVE) (1917), THINK (REFLECT) (1917)                                                                                 \\\hline
 SMELL                            &           1608 &              2 & SMELL (PERCEIVE) (804), SMELL (STINK) (804)                                                                                    \\\hline
 BOIL                             &            338 &              1 & BOIL (OF LIQUID) (338)                                                                                                         \\\hline
 BURN                             &           5012 &              2 & BURN (SOMETHING) (2506), BURNING (2506)                                                                                        \\\hline
 KNOW                             &            689 &              1 & KNOW (SOMETHING) (689)                                                                                                         \\\hline
 EAGLE OR HAWK                    &            382 &              2 & EAGLE (191), HAWK (191)                                                                                                        \\\hline
 ARM OR HAND                      &            720 &              2 & ARM (360), HAND (360)                                                                                                          \\\hline
 FOOT OR LEG                      &           2340 &              2 & FOOT (1170), LEG (1170)                                                                                                        \\\hline
 FLESH OR MEAT                    &           2852 &              2 & FLESH (1426), MEAT (1426)                                                                                                      \\\hline
 PERSPIRE OR SWEAT                &            996 &              2 & SWEAT (PERSPIRE) (498), SWEAT (SUBSTANCE) (498)                                                                                \\\hline
 THIN (OF HAIR AND THREAD)        &             34 &              1 & THIN (OF SHAPE OF OBJECT) (34)                                                                                                 \\\hline
 RAINING OR RAIN                  &           1086 &              2 & RAIN (PRECIPITATION) (543), RAIN (RAINING) (543)                                                                               \\\hline
 BLACK OR DARK                    &            204 &              2 & BLACK (102), DARK (102)                                                                                                        \\\hline
 EARTH OR LAND                    &            402 &              2 & EARTH (SOIL) (201), LAND (201)                                                                                                 \\\hline
 TURN                             &           2620 &              2 & TURN (SOMETHING) (1310), TURN AROUND (1310)                                                                                    \\\hline
 BELLY OR STOMACH                 &             70 &              2 & BELLY (35), STOMACH (35)                                                                                                       \\\hline
 FINGER OR TOE                    &              4 &              2 & FINGER (2), TOE (2)                                                                                                            \\\hline
 WE TWO (INCLUSIVE)               &            302 &              1 & WE TWO (302)                                                                                                                   \\\hline
 HOT OR WARM                      &            274 &              2 & HOT (137), WARM (137)                                                                                                          \\\hline
 SHY OR ASHAMED                   &            607 &              1 & SHY (607)                                                                                                                      \\\hline
 NO OR NOT                        &           2190 &              2 & NO (1095), NOT (1095)                                                                                                          \\\hline
 CLAW OR NAIL                     &            759 &              3 & CLAW (253), FINGERNAIL (253), TOENAIL (253)                                                                                    \\\hline
 BLUE OR GREEN                    &             58 &              2 & BLUE (29), GREEN (29)                                                                                                          \\\hline
 BAD OR EVIL                      &           1344 &              2 & BAD (672), EVIL (672)                                                                                                          \\\hline
 THATCH OR ROOF                   &           1408 &              2 & ROOF (704), THATCH (704)                                                                                                       \\\hline
 PAINFUL OR SICK                  &           1954 &              2 & PAINFUL (977), SICK (977)                                                                                                      \\\hline
 DREAMING OR DREAM                &            514 &              2 & DREAM (257), DREAM (SOMETHING) (257)                                                                                           \\\hline
 LARGE WILD HERBIVORE             &            132 &              1 & DEER (132)                                                                                                                     \\
\hline
\end{longtable}

\end{document}